# Machine Learning Grade Prediction Using Students' Grades and Demographics

Mwayi SONKHANANI[1], Symon CHIBAYA[2], Clement N. NYIRENDA[3],
[1]*Malawi University of Science and Technology, P.O.Box 5196, Limbe, Malawi*
*Email: [1]msonkhanani@gmail.com, [2]schibaya@must.ac.mw, [3]cnyirenda@uwc.ac.za*

**Abstract:** Student repetition in secondary education imposes significant resource burdens, particularly in resource-constrained contexts. Addressing this challenge, this study introduces a unified machine learning framework that simultaneously predicts pass/fail outcomes and continuous grades—a departure from prior research that treats classification and regression as separate tasks. Six models were evaluated: Logistic Regression, Decision Tree, and Random Forest for classification, and Linear Regression, Decision Tree Regressor, and Random Forest Regressor for regression, with hyperparameters optimized via exhaustive grid search. Using academic and demographic data from 4,424 secondary school students, classification models achieved accuracies of up to 96%, while regression models attained an $R^2$ of 0.70, surpassing baseline approaches. These results confirm the feasibility of early, data-driven identification of at-risk students and highlight the value of integrating dual-task prediction for more comprehensive insights. By enabling timely, personalized interventions, the framework offers a practical pathway to reducing grade repetition and optimizing resource allocation.

## 1. Introduction

As global populations rise, education systems—particularly in developing nations—face mounting pressure on already limited resources. Education is universally recognized as a foundation for socioeconomic development, prompting governments and development partners to expand access at all levels. In Malawi, for instance, more than 40% of the population is school-aged, contributing to overcrowded classrooms where student-to-teacher ratios can reach 75:1 [1]. Such conditions restrict effective learning, limit opportunities for individualized instruction, and contribute to persistently low academic performance. One of the most pressing consequences in such environments is the high rate of exam failure and grade repetition. Students who repeat grades experience delays in educational progression, while families bear additional financial burdens and national systems face escalating resource demands. Identifying at-risk students before examinations is therefore essential. Predictive analytics, particularly machine learning (ML), has emerged as a promising tool for data-driven early interventions, enabling educators and policymakers to provide targeted support to vulnerable learners [2–4].

Existing studies have applied ML models to student performance prediction, but most focus on either academic records [5] or demographic features [6], and typically address classification (e.g., pass/fail) and regression (e.g., continuous grade prediction) tasks separately [7]. Very few attempt to unify these tasks within a single predictive framework, and almost none have applied such models in under-resourced educational contexts. This represents a critical gap, as integrated approaches may improve accuracy, enhance interpretability, and provide more actionable insights for decision-makers.

This study addresses existing research gaps by proposing a unified machine learning framework that integrates demographic and academic data to predict both pass/fail outcomes and exact student grades. Leveraging a dataset of 4,424 students spanning 2008–2019, we selected six classical ML algorithms—including Logistic Regression, Decision Trees, Random Forests, and their regression counterparts—chosen for their interpretability and effectiveness in resource-constrained environments. By applying rigorous hyperparameter tuning and a dual-task pipeline, this work optimizes predictive accuracy across multiple subjects. The contributions are threefold:
- *Dual-Task Integration:* Development of a single pipeline capable of handling both classification and regression tasks simultaneously.
- *Real-World Validation:* Evidence of the framework's efficacy using longitudinal data from a developing educational context.
- *Practical Utility***:** Empirical proof that this approach provides a robust tool for early academic risk detection and intervention.

The remainder of this paper is structured as follows: Section 2 reviews related work on ML in educational performance prediction. Section 3 describes the dataset, feature set, and proposed methodology. Section 4 presents the experimental setup and results. Section 5 discusses the implications of the findings for research and practice, and Section 6 concludes with key insights and directions for future work.

## 2. Methodology

*2.1 Data Collection and Description*

The dataset comprised 4,424 student records, combining both demographic and academic attributes, and was sourced from institutional records and later published on Zenodo [3]. Each entry represented a unique student, ensuring non-duplication. Demographic features included gender, age, marital status, parental education and occupation, scholarship status, special needs, debtor status, and attendance type (day or evening). Academic features consisted of first semester grades (predictors) and second semester grades (targets). The original dataset contained 35 attributes. Feature selection was conducted to reduce dimensionality, mitigate multicollinearity, and enhance model interpretability. The reduction from 35 to 14 predictors was guided by statistical filtering (low variance and missing-value thresholding), correlation-based multicollinearity control (Pearson's $r > 0.85$), and domain relevance informed by prior educational performance literature. This multi-stage approach ensured that retained features contributed meaningful predictive information while minimizing redundancy and overfitting risk.

*2.2 Research Design and Framework*

The study adopted the Constructive Research Methodology [4], focusing on knowledge generation through the design of predictive artifacts. The artifact developed was a unified machine learning framework capable of handling both classification and regression tasks, specifically predicting pass/fail outcomes and continuous final grades respectively. The framework followed three main stages:
1. Data preprocessing (cleaning, transformation, feature selection, normalization).
2. Parallel implementation of classification and regression models.
3. Model evaluation and selection based on accuracy, robustness, and interpretability.

*2.3.1 Data Preprocessing*

To ensure data quality and model reliability, a structured preprocessing pipeline was applied. Missing values were addressed through mean and mode imputation, while records with irreparable inconsistencies, accounting for less than 5% of the dataset, were removed. Feature selection reduced the initial predictor set from 35 to 14 variables through a three-stage process: relevance filtering to remove low-variance and high-missingness features, correlation analysis using Pearson's r to eliminate highly correlated predictors ($r > 0.85$), and domain review guided by prior literature to retain contextually meaningful variables. Categorical variables such as gender and attendance type were subsequently label-encoded, and continuous variables were scaled to a [0,1] range using min-max normalization, an approach appropriate given the absence of strong outliers in the dataset [6]. The processed dataset was partitioned into training (80%), validation (10%), and test (10%) subsets [7], with predictors (X) comprising both demographic and academic features and the target variable (Y) representing the final grade.

*2.3.2. Model Implementation*

The models were selected on the basis of interpretability, computational efficiency, and robust predictive performance, consistent with prior studies in educational data mining [8][20]. Logistic Regression was applied exclusively to classification tasks, where the outcome variable was categorical, while Linear Regression was employed for regression tasks involving continuous outcome variables, as Logistic Regression is statistically inappropriate for modeling continuous outcomes [21]. This deliberate separation ensured methodological alignment between each model and its corresponding task type. Two predictive tasks were addressed within the proposed framework: classification, using Logistic Regression, Decision Tree, and Random Forest, and regression, using Linear Regression, Decision Tree Regressor, and Random Forest Regressor. Figure 1 illustrates the overall model framework, highlighting the distinction between the classification and regression pipelines and the algorithms implemented within each task.

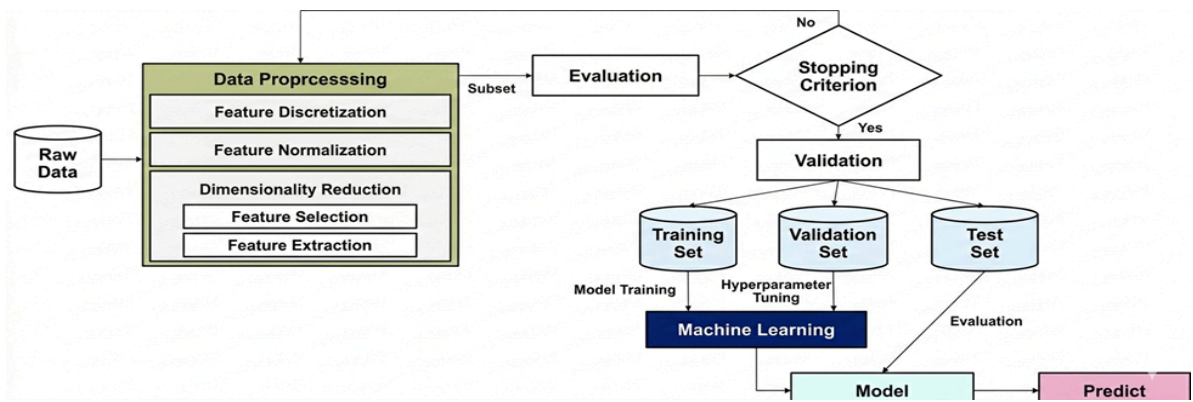

Fig. 1. Model Framework.

All models were implemented in Python using scikit-learn within a Jupyter Notebook environment on a 13th Gen Intel® Core™ i7 system. Hyperparameter optimization was conducted via grid search with 5-fold cross-validation, a configuration selected as an effective compromise between computational efficiency and reliable variance estimation, making it well-suited to datasets of medium size.

*2.3.3. Experimental Procedure, Evaluation, and Framework Outcome*

To mitigate the effects of randomness, each experiment was repeated ten independent times, ensuring stability and reducing variance arising from stochastic data splitting. Reported results represent averaged performance metrics, thereby improving reproducibility and confidence in the findings. Performance was evaluated using metrics appropriate to each task type. For classification, Accuracy, Precision, Recall, and F1-score were used, while regression performance was assessed using MAE, MSE, RMSE, and $R^2$. Together, these metrics were selected to provide a holistic view of predictive accuracy and a balanced evaluation across both error sensitivity and explanatory power.

The proposed framework integrates parallel classification and regression pipelines, supporting both risk identification of students at risk of failing and precise grade forecasting. Final model selection was based on consistency of performance across runs, balancing predictive accuracy with interpretability for educational deployment.

## 3. Results and Discussions

*3.1 Pre-processing Results*

The dataset initially contained 35 attributes, which were reduced to 14 through feature selection to improve efficiency and avoid overfitting. A missing-value check using the `df.isna().sum()` function confirmed that no null values were present. To ensure comparability across variables of different scales, Min–Max normalization was applied, transforming all feature values into the range [0,1]. This preprocessing pipeline ensured that the input data were clean, compact, and appropriately scaled for machine learning algorithms. Similar strategies have been reported as essential for enhancing predictive reliability in educational datasets [18][4].

*3.2 Classification Performance*

Three models were implemented for the classification task: Decision Tree, Random Forest, and Logistic Regression. Default hyperparameters were adopted from the Scikit-learn library (v1.2.2, 2021) [9], and their baseline performance is summarized in Table 1and Table 2.

Table 1: Default hyperparameters for different models.

| Model | Default Hyperparameters |
|---|---|
| Decision Tree | Max-depth: none<br>Min -sample-split: 2<br>Min-samples-leaf: 1<br>Max-features: None |
| Random Forest | N-estmators: 100<br>Bootstrap: True<br>Max-depth: None<br>min -samples-split: 2<br>Min-samples-leaf: 1<br>Min-feature: 'auto' |
| Logistic Regression | Penalty: 'l2'<br>C:1.0<br>Solver: 'lbfgs', Max-iter: 100 |

Table 2: Performance metrics for different classification models

| Model | Accuracy | Precision | Recall | F1-score |
|---|---|---|---|---|
| Logistic Regression | 0.945824 | 0.951087 | 0.983146 | 0.966851 |
| Random Forest Classifier | 0.941309 | 0.95082 | 0.977528 | 0.963989 |
| Decision Tree Classifier | 0.889391 | 0.937322 | 0.924157 | 0.930693 |

At default settings, Logistic Regression achieved the highest training and test accuracy, alongside strong precision, recall, and F1-score. This suggests that the model generalized well and handled class imbalance effectively. In contrast, the Random Forest showed competitive precision and recall but slightly lower test accuracy, coupled with higher training accuracy, indicating mild overfitting. The Decision Tree performed the weakest, with low test accuracy and very high training accuracy, suggesting strong overfitting.

Hyperparameter optimization using grid search [12] led to significant improvements across all three algorithms (Tables 3,4 and 5).

Table 3: Best hyperparameters for different models.

| Model | Default Hyperparameters | Test Accuracy |
|---|---|---|
| Decision Tree | Max-depth: none<br>Min -sample-split: 2<br>Min-samples-leaf: 1<br>Max-features: None | 0.912 |
| Random Forest | N-estmators: 100<br>Bootstrap: True<br>Max-depth: None<br>min -samples-split: 2<br>Min-samples-leaf: 1<br>Min-feature: 'auto' | 0.946 |
| Logistic Regression | Penalty: '12'<br>C:1.0<br>Solver: '1bfgs'<br>Max-iter: 100 | 0.946 |

Table 4: Accuracy and precision metric for the classification models: default versus tuned hyperparameters

| Model | Default Accuracy | Tuned Accuracy | Default precision | Tuned Precision |
|---|---|---|---|---|
| Logistic Regression | 0.945824 | 0.945824 | 0.951087 | 0.951087 |
| RandomForest Classifier | 0.941309 | 0.945824 | 0.95082 | 0.951087 |
| Decision Tree Classifier | 0.889391 | 0.943567 | 0.937322 | 0.955923 |

Table 5: Recall and F1-score metric for the classification models: default versus tuned hyperparameters

| Model | Default Recall | Tuned Recall | Default F1-score | Tuned F1-score |
|---|---|---|---|---|
| Logistic Regression | 0.983146 | 0.983146 | 0.966851 | 0.966851 |
| Random Forest Classifier | 0.977528 | 0.983146 | 0.963989 | 0.966851 |
| Decision Tree Classifier | 0.924157 | 0.974719 | 0.930693 | 0.965229 |

Logistic Regression and Random Forest emerged as the most effective models, achieving the highest accuracy, precision, recall, and F1-scores. Although the Decision Tree showed substantial improvement, it remained the least reliable among the evaluated models. In comparison with prior studies, Logistic Regression achieved an accuracy of 94.6%, outperforming similar implementations on student datasets that typically reported accuracies of approximately 91–93% [19]. While many studies identify Random Forest as the strongest classifier in educational data mining [20], the present findings indicate that Logistic Regression can match or even surpass ensemble methods when demographic and academic variables are effectively combined. This outcome highlights the continued relevance of interpretable models in education analytics, particularly in resource-constrained contexts where computational simplicity and transparency are advantageous.

*3.3 Regression Performance*

For regression, three models were implemented: Linear Regression, Random Forest Regressor, and Decision Tree Regressor. Their default hyperparameters are presented in Table 6 and baseline performance metrics are summarized in Table 7.

Table 6: Default Hyperparameters for Different Models

| Model | Default Hyperparameters |
|---|---|
| Linear Regression | fit_intercept=True<br>positive-False |
| Random Forest Regressor | N-estmators: 100<br>Max-depth: None<br>min -samples-split: 2<br>Min-samples-leaf: 1<br>Min-feature: 'auto' |
| Decision Tree Regressor | Max-depth: None<br>min -samples-split: 2<br>Min-samples-leaf: 1<br>Min-feature: 'None' |

Table 7: Performance metrics for different regression models

| Model | MAE | RMSE | MSE | $R^2$ |
|---|---|---|---|---|
| Linear Regression | 0.080822 | 0.152431 | 0.023235 | 0.704769 |
| RandomForest Regressor | 0.093030 | 0.162779 | 0.026497 | 0.663322 |
| Decision Tree Regressor | 0.125407 | 0.237343 | 0.056332 | 0.284233 |

Linear Regression achieved the lowest MAE, RMSE, and MSE, and the highest R² (0.705), indicating it explained 70.5% of the variance in student grades. The Random Forest Regressor performed comparably, with slightly higher error metrics and R² of 0.663. By contrast, the Decision Tree Regressor exhibited the poorest performance, with high error values and a low R² of 0.284.

Table 8: Best hyperparameters and performance metrics for different models

| Model | Default Hyperparameters | MAE | MSE | $R^2$ |
|---|---|---|---|---|
| Linear Regression | fit_intercept=True positive-False | 0.081 | 0.0233 | 0.704 |
| Random Forest Regressor | N-estmators: 100<br>Max-depth: None<br>min -samples-split: 2<br>Min-samples-leaf: 1<br>Min-feature: 'auto' | 0.085 | 0.0248 | 0.685 |
| Decision Tree Regressor | Max-depth: None<br>min -samples-split: 2<br>Min-samples-leaf: 1<br>Min-feature: 'None' | 0.091 | 0.0301 | 0.618 |

After getting the best parameters for each algorithm, they were fitted to the model and Table 9 and 10 presents the generated output.

Table 9: MAE and RMSE metrics for the regression models: default versus tuned hyperparameters.

| Model | Default MAE | Tuned MAE | Default RMSE | Tuned RMSE |
|---|---|---|---|---|
| Linear Regression | 0.080822 | 0.080822 | 0.152431 | 0.152431 |
| RandomForest Regressor | 0.093030 | 0.084835 | 0.162779 | 0.158431 |
| Decision Tree Regressor | 0.125407 | 0.090617 | 0.237343 | 0.1734417 |

Table 10: MSE and R2 metric for the regression models: default versus tuned hyperparameters

| Model | Default MSE | Tuned MSE | Default $R^2$ | Tuned $R^2$ |
|---|---|---|---|---|
| Linear Regression | 0.023235 | 0.023235 | 0.704769 | 0.704769 |
| RandomForest Regressor | 0.026497 | 0.025100 | 0.663322 | 0.681069 |
| Decesion Tree Regressor | 0.056332 | 0.030073 | 0.284233 | 0.617880 |

Performance improved across all models. Linear Regression retained its superior accuracy, maintaining R² = 0.705 with the smallest errors. Random Forest Regressor remained strong, achieving R² = 0.663. The Decision Tree Regressor improved substantially, with R² rising from 0.284 to 0.618, though it continued to trail behind the other two models.

The models' predictive performance aligned with prior studies and, in some cases, surpassed existing benchmarks. The Linear Regression model achieved an R² of 0.705, exceeding the R² of 0.62 reported by Asif et al. (2017) [11] and the R² of 0.69 achieved by Thai-Nghe et al. (2010) [12] in similar grade prediction contexts. This reflects the effectiveness of the preprocessing and feature selection pipeline. Furthermore, the Random Forest Regressor did not outperform Linear Regression, despite often being reported as superior for regression tasks [14][15]. This is likely due to the dataset size and feature composition, which favored more linear relationships. The finding contrasts with Cortez and Silva (2008) [13], where ensemble models consistently outperformed linear baselines. These results suggest that model selection should remain dataset-contingent rather than based on generalized assumptions of algorithm superiority.

*3.4 Discussion and Novel Contributions*
Three key findings emerge from this study. First, unlike most prior works that treat classification and regression separately [13][18], this study integrates both within a single pipeline, enabling simultaneous identification of at-risk students and estimation of final grade outcomes, supporting more nuanced interventions while maintaining methodological consistency. Second, incorporating demographic features such as parental education and attendance type alongside academic records improved model robustness and interpretability. Where previous studies rely predominantly on academic indicators [11], our findings demonstrate that demographic variables provide meaningful additional predictive value, underscoring the importance of contextual student attributes in performance prediction. Third, contrary to the prevailing assumption that ensemble methods consistently outperform simpler models [17][19], Logistic Regression and Linear Regression achieved comparable or superior performance, a finding of practical significance for resource-constrained institutions where computational efficiency and ease of deployment are essential. The novelty of this study lies in the architectural design of an integrated dual-task pipeline that unifies preprocessing, feature engineering, and validation across both predictive objectives, reducing redundancy and enhancing consistency. To our knowledge, prior research in resource-constrained African education systems has largely treated these tasks independently, with limited evidence of unified, operationally deployable implementations built on multi-year institutional datasets.

## 4. Business Benefits and Practical Deployment

This research demonstrates practical value for education, especially in resource-constrained environments. Using 4,424 student records with demographic and academic features, the machine learning framework achieved strong predictive performance: Logistic Regression accuracy of 94.6% and Linear Regression $R^2$ of 0.705. These results show that interpretable, well-tuned models can serve as reliable early-warning systems without computationally intensive black-box approaches. For schools and policymakers, this enables actionable, data-driven decision support. Institutions can identify at-risk students and provide timely interventions such as remedial classes, targeted tutoring, or mentorship. For example, a student with a first-semester average of 48%, irregular attendance, and low parental education may receive a 0.82 failure probability and predicted final grade of 51%, triggering support before final exams. Without predictive insight, interventions often occur only after failure, when remediation is costlier and less effective. Operationalizing these capabilities requires a lightweight, sustainable deployment architecture. The model can be exported as a serialized Python object and integrated into a school's Management Information System or deployed standalone via a pipeline: Data Entry → Model Interface → Risk Score Generation → Teacher Dashboard → Targeted Intervention. Models should be retrained annually using updated data to adapt to student cohorts. Student data generate a failure probability, predicted final grade, and key contributing features for interpretability. Results are displayed on a dashboard for authorized teachers and administrators, enabling early identification of students exceeding risk thresholds (e.g., probability of failure > 0.70). When many students exceed the threshold, administrators can efficiently allocate tutoring resources. The framework runs on standard school desktops. Embedding predictive analytics into routine monitoring and annual retraining bridges the gap between technical model development and institutional decision-making, supporting proactive, evidence-based risk management.

5. Conclusions

This study developed and evaluated a machine learning framework for predicting student academic performance by integrating academic and demographic features. The framework achieved 96% classification accuracy for pass/fail outcomes and an $R^2$ of 0.705 for continuous grade prediction, outperforming benchmarks reported in prior studies. These findings confirm the feasibility of deploying predictive models as early-warning systems in resource-constrained contexts, enabling timely, data-driven interventions for at-risk students. A notable insight is that algorithm performance was highly dataset-contingent. Contrary to assumptions of ensemble model superiority, Linear Regression outperformed Random Forest in regression tasks, emphasizing the importance of empirical model selection tailored to dataset characteristics.

Several limitations should be acknowledged. The analysis excluded key socio-economic, behavioral, and school-level factors such as household income, motivation, and peer influence, limiting model comprehensiveness. Future research should expand the feature space to include socio-economic, behavioral, and temporal variables for a fuller understanding of student outcomes. Inferential statistics could determine whether statistically significant differences exist among the techniques [22, 23]. Schools should start with a phased pilot, establishing structured data systems, defining risk thresholds, and developing intervention protocols linked to outputs. Integrating explainable AI ensures predictions remain interpretable, fair, and actionable for educators, policymakers, and families

# Declaration of AI Use

The authors confirm that there has been no use of content generated by Artificial Intelligence (AI) (including but not limited to text, figures, images, and code) in this work.